\documentclass[conference]{IEEEtran}

\IEEEoverridecommandlockouts

\usepackage{cite}
\usepackage{amsmath,amssymb,amsfonts}
\usepackage{algorithmic}
\usepackage{graphicx}
\usepackage{textcomp}
\usepackage{xcolor}
\def\BibTeX{{\rm B\kern-.05em{\sc i\kern-.025em b}\kern-.08em
    T\kern-.1667em\lower.7ex\hbox{E}\kern-.125emX}}

\usepackage{float}

\usepackage{booktabs}

\begin{document}

\title{Combining Monte-Carlo Tree Search with Proof-Number Search\\

\author{\IEEEauthorblockN{Elliot Doe, Mark H. M. Winands, Dennis J. N. J. Soemers and Cameron Browne}
\IEEEauthorblockA{\textit{Games Group, Department of Data Science and Knowledge Engineering, Maastricht University} \\
Maastricht, The Netherlands\\ e.doe@student.maastrichtuniversity.nl, \{m.winands, dennis.soemers, cameron.browne\}@maastrichtuniversity.nl}}
}


%

\maketitle

\begin{abstract}
Proof-Number Search (PNS) and Monte-Carlo Tree Search (MCTS) have been successfully applied for decision making in a range of games.  This paper proposes a new approach called PN-MCTS that combines these two tree-search methods by incorporating the concept of proof and disproof numbers into the UCT formula of MCTS.   Experimental results demonstrate that PN-MCTS outperforms basic MCTS in several games including Lines of Action, MiniShogi, Knightthrough, and Awari, achieving win rates up to 94.0\%.
\end{abstract}


\section{Introduction}

In recent years a new paradigm for game-tree search has emerged,
called Monte-Carlo Tree Search (MCTS)~\cite{coulom06,kocsis06b}.
%
 It is a best-first search method guided by the results of Monte-Carlo simulations.  Using the results of previous simulations, the method gradually builds up a game tree in memory and increasingly becomes better at accurately estimating the values of the most promising moves. MCTS has substantially advanced the state of the art in several deterministic game domains \cite{mctssurvey}, in particular
Go \cite{Silver2017mastering}, but other board games include  Amazons~\cite{Lorentz08}, Hex
\cite{arneson10}, Lines of Action \cite{winands10},  and the ones of the  General Game
Playing competition~\cite{bjornsson09}. MCTS has even increased the level of competitive agents in board games with challenging properties such as multi-player (e.g., Chinese Checkers \cite{sturtevant08}) and uncertainty (e.g., Kriegspiel \cite{ciancarini10} and Scotland Yard  \cite{nijssen12tciaig}).


In tactical games, where the main line towards the winning
position is typically narrow with many non-progressing alternatives, MCTS may often lead to an erroneous outcome because the nodes' values in the tree do not converge fast enough to their game-theoretic value. To mitigate this effect, MCTS variants have been proposed that integrate concepts of minimax search \cite{winands08b,winands11,LanctotWPS14,baier2015}.

Another promising direction would be the incorporation of  Proof-Number Search (PNS) \cite{allis94} in MCTS. PNS and its variants \cite{vandenHerik2008} have been proposed to prove endgames faster than traditional minimax. PNS variants  have been successfully applied to a large number of domains including
Chess \cite{breuker}, Othello \cite{nagai}, Shogi \cite{nagai}, Lines of Action (LOA) \cite{Winands04}, Go \cite{Kishimoto05b}, Checkers \cite{schaeffer07a}, Connect6 \cite{Wu10}, and the multi-player game Rolit \cite{Saito10}.  All PNS variants share two features: (1) they are algorithms for solving binary goals, such as proving a win or a loss for a game position, and (2) they rely on the concept of proof and disproof numbers.

This paper proposes a new variant, called PN-MCTS, that combines the strengths of MCTS and PNS with each other. The idea is to incorporate proof and disproof numbers in the UCT mechanism \cite{kocsis06b} of MCTS. To investigate its performance, game-playing experiments are conducted in five two-player board games (i.e., Lines of Action, Gomoku, MiniShogi, Knightthrough, and Awari). 
 

This paper is organized as follows. First, MCTS and PNS are discussed in Sections \ref{sec:MCTS} and \ref{sec:PNS}, respectively. Next, PN-MCTS is proposed in Section \ref{sec:PN-MCTS}. Subsequently, the proposed technique is empirically evaluated in Section \ref{sec:Exp}. Finally, Section \ref{sec:Conc} gives conclusions and an outlook on future research. 


\section{Monte-Carlo Tree Search}\label{sec:MCTS}
\begin{figure}[hb]
    \centerline{\includegraphics[width= \columnwidth]{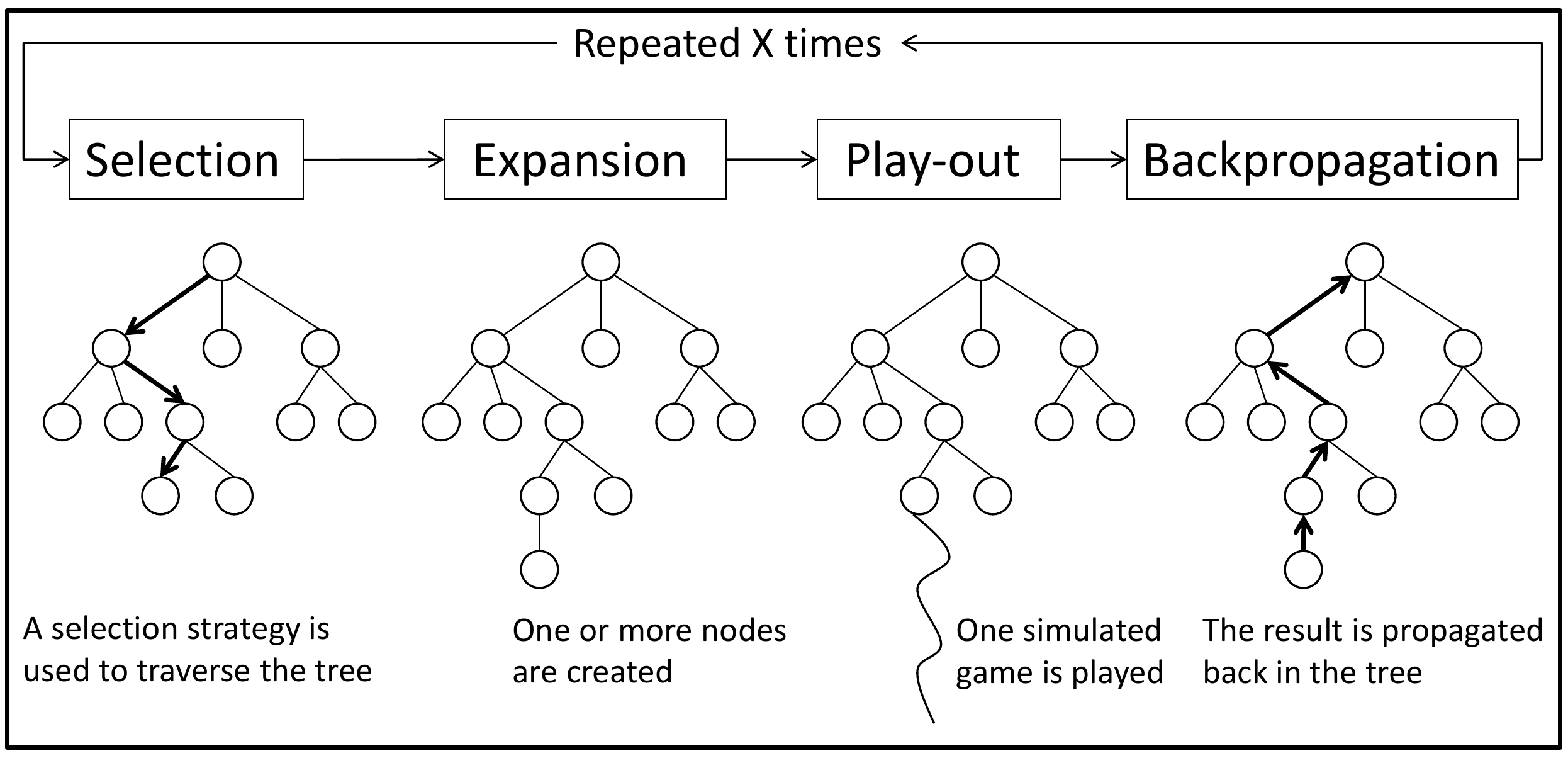}}
    \caption{Outline of Monte-Carlo Tree Search.}
    \label{fig:mcts}
\end{figure}

Monte-Carlo Tree Search (MCTS) \cite{coulom06,kocsis06b} is a best-first search method that does
not require a positional evaluation function. It is based on a
randomized exploration of the search space. Using the results of
previous explorations, the algorithm gradually builds up a game
tree in memory, and increasingly becomes better at accurately
estimating the values of the most promising moves. MCTS consists of four strategic steps, repeated as long as there is
time left \cite{chaslot08}. The steps, outlined in \figurename~\ref{fig:mcts}, are as follows.\\

\textbf{Selection Step.} In the first step, a  child is selected to be searched based on previous gathered information.
The selection step controls the balance between exploitation and exploration. On the
one hand, the task consists of selecting the move that leads
to the best results so far (exploitation). On the other hand, the less promising moves still have to be tried, due to the uncertainty of
the simulations (exploration).

Several \textit{selection strategies} \cite{mctssurvey}  have been suggested for MCTS such as BAST, EXP3, UCB1-Tuned, but the most popular one is based on the UCB1 algorithm \cite{auerfinit02}, called  UCT (\textbf{U}pper \textbf{C}onfidence Bounds applied to
\textbf{T}rees) \cite{kocsis06b}. UCT works as follows. Let $I$ be the set of nodes immediately reachable from the
current node $p$. The selection strategy selects the child $b$ of
node $p$ that satisfies Formula \ref{eq:UCT}:

 \begin{equation}
 \label{eq:UCT}
 \mathit{b\in argmax_{i \in I} \left(v_i + C \times \sqrt{\frac{\ln{n_p}}{n_i}}\right)}
 \end{equation}

\noindent where $v_i$ is the value of the node $i$, $n_i$ is the
visit count of $i$, and $n_p$ is the visit count of $p$. $C$ is a
parameter constant, which can be tuned experimentally (e.g., $C=\sqrt{2}$).
In the case of a tie, the tie is broken randomly. This process is repeated until a node is reached that has not yet fully been expanded.\\

\textbf{Expansion Step.} As previously stated, the selection step continues until a node is reached that has not yet expanded all of its children. Among the children that have not been stored in tree, one is selected uniformly at random. This node $L$ is then added as a  new leaf node and is subsequently investigated.\\

\textbf{Play-out Step.} From the leaf node the play-out step is performed. Moves are selected in self-play until the end of the game is reached. This step might consist of playing plain random moves or -- better -- semi-random moves chosen according to a \textit{simulation strategy}.\\

\textbf{Backpropagation Step.} In the final step, the result \textit{R}
of a play-out $k$ is backpropagated from the leaf node $L$, through the
previously traversed nodes, all the way up to the root. The result is scored positively $(R_k=+1)$ if the game is won, and negatively
  $(R_k=-1)$ if the game is lost. Draws lead to a result $R_k=0$. A \emph{backpropagation strategy} is applied to the  \textit{value} $v_i$
  of a node $i$. Here, it is computed by taking the average of the results of all simulated games made through this node \cite{coulom06},
   i.e., $v_i=(\sum_k R_k ) / n_i$.

\section{Proof-Number Search}\label{sec:PNS}
 Proof-Number Search (PNS) is a best-first search method especially suited for finding the
game-theoretic value in game trees \cite{allis94}. Its aim is to prove
a particular goal. In the context of this paper, the goal is to prove a forced win for the player to move. A tree can have three values:
\textit{true}, \textit{false}, or \textit{unknown}.  In the case of a
forced win, the tree is \textit{proven}  and its value is true. In
the case of a forced loss or draw, the tree is \textit{disproven}
and its value is false. Otherwise, the value of the tree is unknown.
As long as the value of the root is unknown, the most-promising node
is expanded. Just like MCTS, PNS
does not need a domain-dependent heuristic evaluation function to
determine the most-promising node \cite{allis94}. In PNS this
node is usually called the \textit{most-proving} node. PNS
selects the most-proving node using two criteria: (1) the shape of
the search tree (the branching factor of every internal node) and
(2) the values of the leaves. These two criteria enable PNS to
treat game trees with a non-uniform branching factor efficiently.

\begin{figure}[ht]
\centerline{\includegraphics[width=\columnwidth]{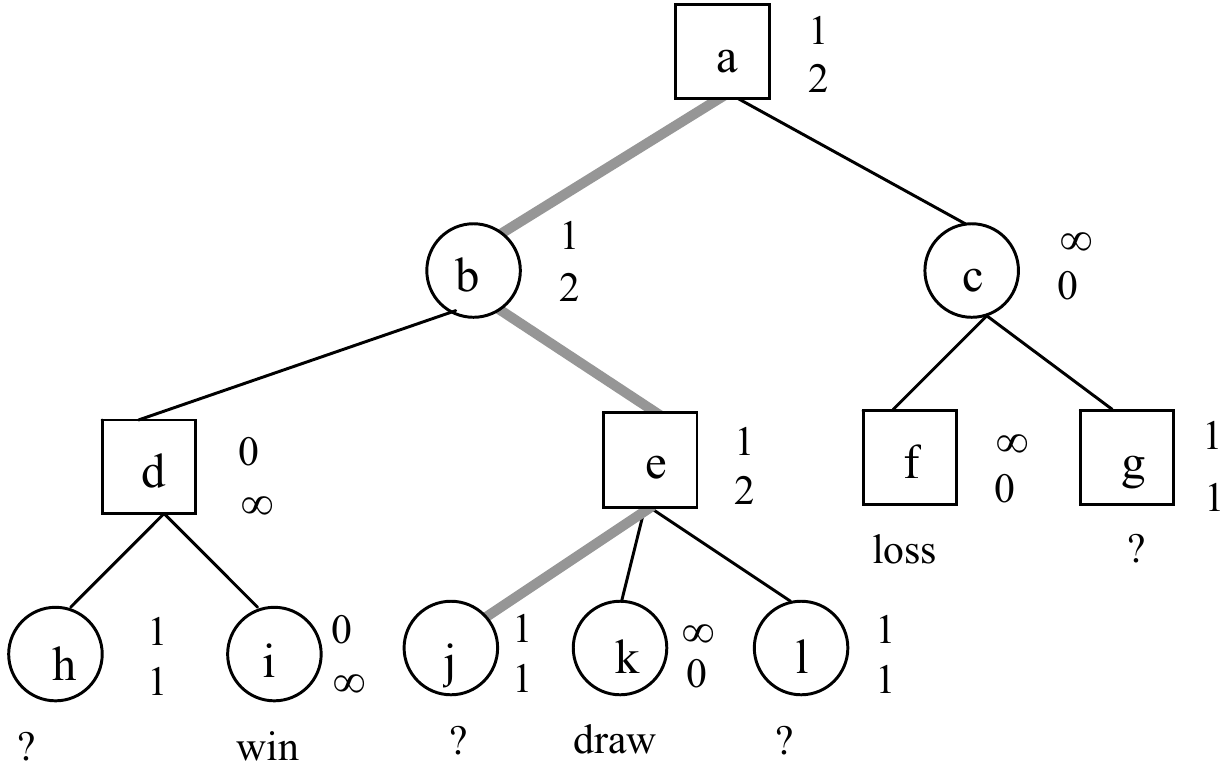} }
\caption{An AND/OR tree with proof and disproof numbers}
 \label{pntree}
\end{figure}

Below we explain PNS on the basis of the AND/OR tree depicted
in Fig. \ref{pntree}, in which a square denotes an OR node, and a
circle denotes an AND node. The numbers to the right of a node
denote the proof number (upper) and disproof
number  (lower). A \textit{proof number}
(\emph{pn}) represents the minimum number of leaf nodes, which have
to be proven in order to prove the node. Analogously, a
\textit{disproof number} (\emph{dpn}) represents the minimum number
of leaf nodes that have to be disproved in order to disprove the
node. Because the goal of the tree is to prove a forced win, winning
nodes are regarded as proven. Therefore, they have $pn=0$ and
$dpn=\infty$ (e.g., node \textit{i}). Lost or drawn
nodes are regarded as disproven (e.g., nodes \textit{f} and
\textit{k}). They have $pn =\infty$ and $dpn=0$.
Unknown leaf nodes have  $pn=1$ and $dpn=1$ (e.g.,
nodes \textit{g}, \textit{h}, \textit{j}, and \textit{l}).  The \emph{pn} of an
internal OR node is equal to the minimum of its children's proof
numbers, since to prove an OR node it suffices to prove one child.
The \emph{dpn} of an internal OR node is equal to the sum of
its children's disproof numbers, since to disprove an OR node all
the children have to be disproven. The  \emph{pn} of an internal AND node is equal to the sum of its children's
proof numbers, since to prove an AND node all the children have to
be proved. The \emph{dpn} of an AND node is equal to the
minimum of its children's disproof numbers, since to disprove an AND
node it suffices to disprove one child. The procedure of selecting the most-proving node to expand next is as
follows. The algorithm starts at the root. Then, at each OR node the child with
the smallest \emph{pn} is selected as successor, and at each AND
node the child with the smallest  \emph{dpn} is selected as
successor. Finally, when a leaf node is reached, it is expanded
(which makes the leaf node an internal node) and the newborn
children are evaluated. This is called
\textit{immediate evaluation}. The selection of the most-proving
node (\textit{j}) in Fig. \ref{pntree} is given by the bold path.

\section{PN-MCTS}\label{sec:PN-MCTS}
To properly determine a feasible approach to the incorporation of  (dis)proof numbers into MCTS, it is first important to consider what information the proof and disproof numbers bring. As explained in the previous section about PNS, a (dis)proof number provides a lower bound for the number of nodes that still have to be (dis)proven to prove the current node. In PNS, this lower bound (together with the lower bound that the disproof number gives)  determines which leaf node would be investigated further. In MCTS,  the selection step has a similar function. In the default  MCTS implementation, this would be the UCT formula (\ref{eq:UCT}). 

Thus, a natural way to combine MCTS and PNS would be to combine these two ways of selecting promising leaf nodes. The approach proposed for this paper is a modification of basic UCT. By adjusting UCT to also use proof and disproof numbers, nothing else about MCTS has to be changed, though the information from the (dis)proof number can still influence the decision making process.

The final consideration then is how to use (dis)proof numbers in UCT. 
The magnitudes of differences amongst (dis)proof numbers technically do not have much meaning. For example, a node with a proof number of 100 is not necessarily ten times worse than a node with a proof number of 10. The node with the proof number of 100 may just have been investigated more often already. The fact that the magnitudes of differences between (dis)proof numbers do not have much meaning makes it difficult to directly use them in the UCT formula.
Instead of using the proof or disproof numbers directly in the formula, this paper proposes that the (dis)proof numbers are used to determine a ranking amongst all the nodes. The ranking is similar to the one of PNS as explained in Section \ref{sec:PNS}. At an OR node the child node with the lowest proof number would get the best ranking because that is the node that would be selected in regular PNS. 
For example, if there are 30 child nodes, the one that would be picked according to PNS gets a rank of 1. 
At an AND-node, the node with the lowest disproof number is picked. The next best ranking node would be the node PNS would pick if the original best ranking node was not an option (so the second lowest (dis)proof number would get a ranking of 2, whereas the worst option would get 30 for this example). Ties are awarded the same rank. Finally, this rank can then be normalized to be in the range of [0, 1]. Normalization allows the resulting value to be in a range that is similar to the values that might come out of the exploitation or exploration parts of the UCT formula. To normalize, the ranking (called $pnRank$ in the formula) is divided by the highest rank of any of the children. Finally, to control the influence of the addition, a parameter is added (called $C_{pn}$ in the formula).

Putting all that together gives the following adjusted UCT formula, referred to as the UCT-PN formula:

\begin{scriptsize}
\begin{equation} \label{UCT-PN_formula}
 \mathit{b\in argmax_{i \in I} \left(v_i + C \sqrt{\frac{\ln{n_p}}{n_i}} + C_{pn} \left( 1 - \frac{pnRank_{i}}{argmax_{j \in I} (pnRank_{j})} \right) \right)}
\end{equation}
\end{scriptsize}

\noindent where $C_{pn}$ is the PN-Parameter that can be adjusted, $pnRank$ is the rank of a specific node (lowest rank is best node according to PNS), and $argmax_{j \in I} (pnRank_{j})$ is the highest (and thus worst) rank of any child node. The rest of the variables are the same as the regular UCT Formula \ref{eq:UCT}. This paper uses the term PN-MCTS to refer to any MCTS variant that uses the UCT-PN formula instead of the base UCT formula for its selection step.


\section{Experiments}\label{sec:Exp}
This section outlines the experiments that have been conducted, the experimental environment and the specific setups. All of the the experiments have been  conducted in the Ludii General Game System \cite{Piette2020Ludii}, and have been run on a Intel(R) Core(TM) i7-10750H CPU  with 16 GB of RAM.                             
First, in Subsection \ref{ludii} the Ludii General Game System is outlined and the choice for this system is explained. Next, in Subsection \ref{domains} the test domains are described. Subsequently, in Subsection \ref{expsetup} the specific experimental setup is explained. Finally, in Subsection \ref{results} the results of the experiments are shown and discussed. 

\subsection{Ludii General Game System}\label{ludii}
The Ludii General Game System is a general game playing framework, which provides an environment for developers to test their implementation of general game playing agents. The system includes over 1,000 games described in its general game description language, and implementations of various standard algorithms and enhancements (such as several variants of MCTS).
It has a single, unified API for the development of intelligent agents, based on a forward model (with functions to generate lists of legal actions, generate successor states, and so on) and standardized state and action representations.
Ludii has been demonstrated \cite{Piette2020Ludii} to process games faster than the previous state-of-the-art general game playing framework based on Stanford's general game description language \cite{Love_2008_GDL}, which is important for the playing strength of tree search algorithms such as MCTS.

\subsection{Game Domains}\label{domains}
Two-player adversarial games are well suited to PNS as it structures its knowledge as AND/OR-trees. The list of games that fulfill this condition is still rather large. 
To narrow the list down even more, only domains in which both MCTS and PNS have shown to be effective are considered.
If either MCTS or PNS does not perform well in a domain, the combination of the two will probably not be very effective.
From the remaining list of games that fits the requirements and desirable qualities, five games are chosen: Lines of Action, Awari, MiniShogi, Gomoku, and Knightthrough. Lines of Action (LOA) is the primary domain for most of the experiments. The other games are used in two of the experiments. 
Each of the games is briefly described below. 

\subsubsection{Lines of Action}

The rules of Lines of Action (LOA) are as follows \cite{sackson69}. It is played on an
8$\times$8 board by two sides, Black and White. Each side has twelve
pieces at its disposal. The black pieces are placed along the top and bottom rows of the board, while the white pieces are placed in
the left- and right-most files of the board. The players alternately move a piece, starting with Black. A piece moves in a straight line, exactly as many squares as there are pieces of either color anywhere along the line of movement. A player may jump over its own pieces, but not the
opponent's, although opposing pieces are captured by landing on them.
The goal of the players is to be the first  to create a configuration on the board in which all own pieces are connected in
one unit. The connections within the unit may be either orthogonal or diagonal.

There are two main reasons why LOA was chosen as the main test domain. Both MCTS and PNS have been extensively tested on LOA \cite{vandenHerik2008,winands10}, and the game board has an adjustable size. The default board is 8$\times$8, but  smaller sizes such as 7$\times$7 can also be used. The advantage of the smaller board sizes is that the game reaches endgame states much quicker. PNS works best in endgame scenarios where game states can be proven or disproven in fewer steps. Thus, by testing on various board sizes, the experiments can test whether PN-MCTS has a better performance when endgame states require fewer steps to be reached.

\subsubsection{Awari}
Awari is a Mancala or sowing game \cite{awari}. Awari is played on a 2$\times$6 board and with counters. The goal of the game is to capture as many counters as possible. To capture counters, a player must end their sow in the opponent's row and in a hole with 2 or 3 counters (including the piece used to sow). Sowing is a process where a player takes all the counters from a hole in their row and deposits them one by one into adjacent holes until none are left. In Awari, sowing goes counter-clockwise. The game is over once none of the holes contain more than 1 counter. The player who captured most counters wins, or in case both players have an equal amount, the game ends in a draw.  Awari is a suitable test domain for PN-MCTS as Mancala variants have been used as testbed for PNS \cite{allis94} and MCTS \cite{LanctotWPS14} in the past.

\subsubsection{MiniShogi}
MiniShogi is a variant of the old Japanese game called Shogi and was invented around 1970 by Shigenobu Kusumoto. The game has various pieces each of which have their own rules for movement. The goal is to capture the opponent's King with these pieces. Pieces can be promoted by moving them towards the opponent's side of the board. Opponent pieces can be captured by moving a piece onto an enemy piece. Captured pieces can re-enter the game as a turn. MiniShogi differs from regular Shogi in the following ways: it is played on a 5$\times$5 board, it has fewer pieces than the original, and features a smaller promotion area. Shogi endgames have  been one of the main test domains of PNS \cite{KishimotoW0S12}, whereas MCTS has become the dominating search technique for this game \cite{SilHub18General}.

\subsubsection{Gomoku}
Gomoku is a connection game \cite{gomoku}. The goal of the game is to make a row of exactly 5 stones of the same color. Gomoku is played on a 15$\times$15 board, with black and white stones. 
Players take turns placing pieces of their color on any empty position of the board.
The game ends once a player places a stone that gives them a line of exactly 5 pieces (lines of more than 5 pieces do not count). Lines may be orthogonal or diagonal. PNS has been applied in Gomoku before \cite{allis96}, and MCTS variants have also been developed for this game \cite{TangZSL16}.

\subsubsection{Knightthrough}
The game of Knightthrough is a variant of Breakthrough \cite{handscomb01}. It is played on an 8$\times$8 board. Each player has 16 pieces in the first two rows of their side of the board (opposing sides). Every piece moves like a knight in chess. This means each piece may move 1 square in one non-diagonal direction and then 2 squares in a perpendicular direction. Pieces may be captured by landing on them. Knights can jump over other pieces (both friendly and opponent). The goal of the game is to reach the opponent's edge of the board (the row furthest from the player) with one of their knights. A player can also win by capturing all opposing pieces. Knightthrough has been used to test MCTS in a general-game-playing context \cite{SironiLW20}. Its original variant Breakthrough has served as a test bed for PNS variants \cite{saffidineJC11}. 

\subsection{Experimental Setup}\label{expsetup}
This subsection explains the setup for the following four experiments.
\subsubsection{Overhead Cost}
This experiment investigates the cost of obtaining — and continuously updating — the proof and disproof numbers. PN-MCTS is constructed as an augmentation of the basic MCTS implementation provided by the Ludii system to ensure that any difference in performance is solely due to the proposed enhancement. To obtain the cost specifically, PN-MCTS and the original MCTS are both run for a fixed amount of time from the initial positions of several different games. The number of simulations that the two search techniques are able to perform are tracked and noted.

\subsubsection{PN-Parameter}
The second experiment tests the performance of PN-MCTS against the original MCTS.  for $C_{pn} \in \{0.0, 0.1, 0.5, 1.0, 2.0, 5.0,  10^6\}$ are tested, each with 250 games against the base MCTS. One of the tested $C_{pn}$ configurations is set to an arbitrarily high number, such that the PN-MCTS performs practically the same as a regular PNS would, with the exception that ties in proof and disproof number are broken by MCTS instead of randomly. The experiments are executed with 1 second per turn. For both agents the $C$ parameter is set to $\sqrt{2}$. This experiment is conducted in the game of LOA. The boards on which the experiments happen are 7$\times$7 and 8$\times$8. By playing on two board sizes, results can be compared to determine if larger search spaces, longer games and later endgame situations have different trends in which parameters are best for the PN-MCTS.

\subsubsection{Time Settings}
This experiment explores what effect a longer or shorter time setting has on the win rate of PN-MCTS. The default MCTS is given the same amount of time per turn. The performance of MCTS becomes significantly better with more time. If only the PN-MCTS was given more time, it would not be possible to say much about the performance of the enhancements. This way the win rate is relative to an MCTS on equal ground. These experiments are conducted with $C_{pn}$ set to 1 for the PN-MCTS and both agents have $C = \sqrt{2}$. As in the previous experiment, for each data point  250 games were played. The domain for this experiment is also LOA on the 7$\times$7 and 8$\times$8 boards.


\subsubsection{Other Domains}
Finally, the last series of experiments investigate the performance of PN-MCTS in domains other than LOA. As described in Subsection \ref{domains}, these other domains are Awari, MiniShogi, Gomoku, and Knightthrough. In each of these games, a subset of the previous experiments is executed. Specifically, a limited version of the time settings experiment is conducted for three time configurations (0.1s, short turns; 1.0s, regular turns; 2.0s, longer turns). LOA with the default board size of 8$\times$8 is also included in the experiment for the sake of comparison. The PN-MCTS parameters for all domains are set as follows:  $C_{pn}=1$, and $C=\sqrt{2}$.


\subsection{Results}\label{results}
In this subsection, the results of the experiments are shown and discussed.  Every result (except for the overhead-cost experiment) has a 95\%-confidence interval behind it in brackets, and the agents play both colors equally.

\subsubsection{Overhead Cost}
In Table \ref{tab:simsPerSecond} the average number of simulations per second is given for PN-MCTS and MCTS, as well as the ratio between these two numbers, for five games. These numbers were measured over a period of 30 seconds for Knightthrough, and 100 seconds for the remaining games. A shorter duration was used for Knightthrough because, due to its large branching factor and relatively high baseline number of simulations per second, PN-MCTS tended to run out of memory  at higher time settings. The increased memory usage is because the implementation of PN-MCTS fully expands all children of nodes at once---allowing for the computation of (dis)proof numbers and rankings for all children---whereas the standard MCTS only expands one child per simulation.
The table reveals that PN-MCTS has a relatively mild overhead in most games (ratios relatively close to $1.0$), with the most notable drop in performance being in Knightthrough. 
While this number does give an indication of the overhead, it should also be noted that the overhead here might be higher than in usual experiments; this experiment measured overhead over 100 seconds (or 30 for Knightthrough), whereas other experiments run tree searches only for up to 5 seconds. 
The (dis)proof numbers have to be recalculated on backpropagation, thus the larger the tree, the higher the backpropagation has to go and the higher the processing cost. However, the (dis)proof numbers for ancestors of a leaf node are only recalculated if the change in (dis)proof number has an effect on the (dis)proof numbers of the ancestors. So even if the game tree goes quite deep (such as in this experiment), the overhead cost should not be much higher than in usual cases. Still, the overhead cost will be different every turn depending on the current board state. 

\begin{table}[H]
\begin{center}
    \caption{Average number of simulations per second}
    \label{tab:simsPerSecond}
    \begin{tabular}{@{}lrrr@{}}
    \toprule
    Game & PN-MCTS & MCTS & Ratio \\
    \midrule
    LOA 8$\times$8 & 172.84 & 188.95 & 0.91 \\
    LOA 7$\times$7 & 312.89 & 341.34 & 0.92 \\
    Awari & 898.82 & 913.69 & 0.98 \\
    MiniShogi & 83.40 & 89.51 & 0.93 \\
    Knightthrough & 1104.23 & 1342.30 & 0.82 \\
    \bottomrule
    \end{tabular}
\end{center}
\end{table}


\subsubsection{PN-Parameter}
\begin{figure}[t]
    \centering
    \includegraphics[width= \columnwidth]{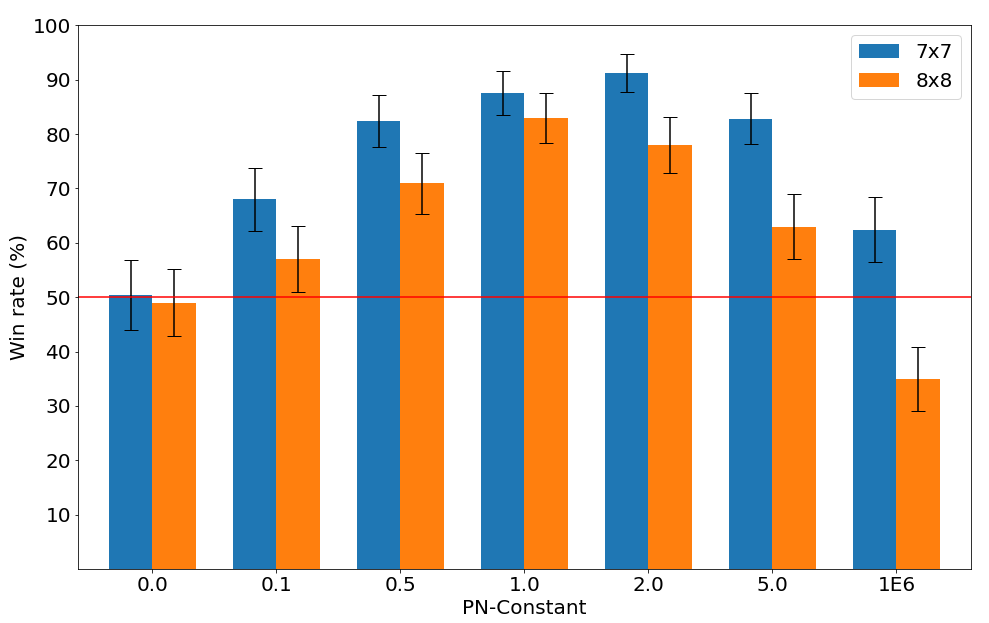}
    \caption{PN-Parameter Experiments: PN-MCTS against MCTS for 7$\times$7 and \& 8$\times$8 LOA}
    \label{fig:pnConstantLOAComparison}
\end{figure}

Figure \ref{fig:pnConstantLOAComparison} displays the win rates of PN-MCTS with varying $C_{pn}$ against MCTS. The trend in win rate of PN-MCTS is fairly similar for both board sizes. If $C_{pn}$ nears 0, such as when it is 0.1, the PN-MCTS will function more like the basic MCTS. If $C_{pn}$ was 0, the two agents would be the same and a win rate of approximately 50\% would be expected. Thus it seems that as $C_{pn}$ nears 0, so too does the win rate near 50\%. Another low point for both board sizes is when $C_{pn}$ is $10^6$. When $C_{pn}$ is that high, the agent starts behaving similar to PNS. From the results, it seems that a basic PNS still performs better than basic MCTS on the smaller 7$\times$7 board, but not on the regular 8$\times$8 board. This is expected as PNS is most effective in endgame play. Endgame play is reached sooner on a 7$\times$7 board and thus (MCTS-)PNS performs better more quickly.
As for the highest win rates, on a 7$\times$7 board, a $C_{pn}$ of 2.0 performs best, recording a win rate of 91.2\% over basic MCTS. On the 8$\times$8 board, there is a different optimal setting. A $C_{pn}$ of 1.0 appears to be the optimal one with a win rate of 83.2\%. 

$C_{pn}$ represents the impact that the (dis)proof numbers have on the basic MCTS. So on the smaller board size, where PNS performs better than MCTS, a bigger influence of the proof and disproof number performs better. This is only true to some degree, as when $C_{pn}$ reaches 5.0, the win rate drops as it starts to converge towards the score of a pure PNS. 

To give an idea how a pure PNS would behave on its own without any CPU overhead, we conducted an experiment with $C_{pn}=10^6$ and a fixed number of 1000 simulations per move in  LOA 8$\times$8. Here PN-MCTS won only 9 out 100 games against the regular MCTS. For the same number of simulations, but for $C_{pn}=1$, PN-MCTS won 99 out 100 games. These results validate that PNS on its own is weaker than MCTS, but when combined together in PN-MCTS the algorithm may outperform MCTS.

\subsubsection{Time Settings}

\begin{figure}[H]
    \centering
    \includegraphics[width= \columnwidth]{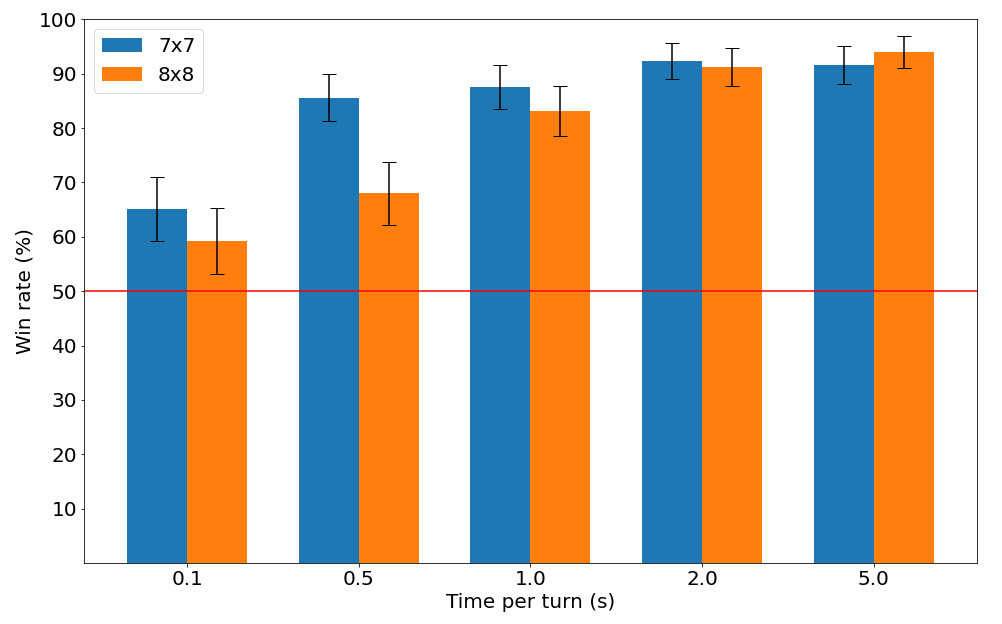}
    \caption{Time Setting Experiments: LOA 7$\times$7 \& 8$\times$8 Compared}
    \label{fig:timeLOAComparison}
\end{figure}

In the next series of experiments, the effect of time on PN-MCTS is measured. 
Figure \ref{fig:timeLOAComparison} shows the performance of PN-MCTS against MCTS for different time settings in LOA 7$\times$7 and 8$\times$8. For each data point  both the PN-MCTS and the basic MCTS have the same time allotted. Just as with the PN-Parameter experiments, the two board sizes show similar trends in their results. Generally, the results indicate that the more time is given to PN-MCTS the higher the win rate, resulting in a 94\% win rate for 5 seconds in  LOA  8$\times$8.  As described before, generally MCTS performs better with more time. This means that any increase in PN-MCTS performance is not only a general improvement but also one relative to the inherent improvement that any enhanced MCTS would see with more time. If this was not the case, the win rate would not go up, and instead stay roughly the same. 

The behavior is likely due to the nature of the enhancement given to PN-MCTS. The PN component of the PN-MCTS has most impact in endgame positions. For the search to reach those positions, time is required. Once those positions  are reached and explored by PN-MCTS, more time does not necessarily improve the decision making much more. The time required for PN-MCTS to reach those endgame positions is shorter on smaller boards. Thus, the aforementioned situation where more time does not necessarily improve the decision making much more is reached sooner too. This is the most likely explanation for the difference in behavior on the different board sizes. Moreover, with enough time PN-MCTS can prove the tree and thus does not need any more time (as proving the tree means PN-MCTS can guarantee a win). 

\subsubsection{Other Domains}
In the final series of experiments, PN-MCTS is tested in four additional domains. Table \ref{tab:domains} shows the performance  of PN-MCTS for different time settings in LOA, MiniShogi, Knightthrough, Awari, and Gomoku. The highest win rate for each domain is printed in bold. The table reveals that in every game except Gomoku, PN-MCTS outperforms MCTS for certain time settings. In some games such as Awari and MiniShogi, PN-MCTS is most effective if the available time is limited (only 0.1 seconds). Due to the limited number of  simulations, the Monte-Carlo evaluations are not very informative, which is mitigated by the  information of the (dis)proof number. In the case, the search is directed to areas where the player has more options than the opponents. Such an implicit mobility feature is correlated with a material advantage. When the search time increases, the Monte-Carlo evaluations become more reliable, and the addition of the (dis)proof number becomes even detrimental.  In other games such as LOA and Knightthrough the opposite seems to be true. With more time, PN-MCTS becomes increasingly stronger than MCTS. 


    

\begin{table}[h!!]
\begin{center}
    \caption{Domain Experiment: Win rate\% of PN-MCTS in various games at various time settings against basic MCTS}
    \label{tab:domains}
    \begin{tabular}{@{}lrrr@{}}
    \toprule
    Game domain & 0.1s & 1.0s & 2.0s\\
    \midrule
    LOA 8$\times$8 & 65.2($\pm$5.90) & 87.6($\pm$4.09) & \textbf{92.4($\pm$3.28)}\\
    MiniShogi & \textbf{86.0($\pm$4.30)} & 76.8($\pm$5.23) & 67.2($\pm$5.82)\\
    Knightthrough & 46.0($\pm$6.18) & 60.4($\pm$6.06) & \textbf{63.6($\pm$5.96)}\\
    Awari & \textbf{61.2($\pm$6.04)} & 49.8($\pm$6.20) & 31.1($\pm$5.74)\\
    Gomoku & \textbf{32.4($\pm$5.80)} & 9.6($\pm$3.65) & 11.2($\pm$3.91)\\
    \bottomrule
    \end{tabular}
\end{center}
\end{table}

\section{Conclusions and Future Research}\label{sec:Conc}
This paper has proposed PN-MCTS, which combines Proof-Number Search and MCTS by adjusting the UCT formula. The ranking of the nodes according to their (dis)proof numbers is used to bias the UCT formula. 
 While there is a computational cost necessary to obtain the proof and disproof numbers, the benefits can outweigh the costs if the domain and choice of parameters are right. The results show that PN-MCTS outperforms  MCTS in LOA, Knightthrough, and MiniShogi. Though for MiniShogi the added benefit seems to diminish when the thinking time increases. For other domains such as Awari and Gomoku, the addition of (dis)proof numbers appears to be detrimental in the long run. One of the reasons could be that PNS can only deal with binary outcomes. In the current PN-MCTS implementation, the (dis)proofnumber is the same for a  draws and a loss, potentially steering the search into the wrong direction. 


There are three main directions for future research. The first one is to test PN-MCTS on more domains and to investigate the reason why it  does (not) work for certain domains.  The second one is to test PN-MCTS with different parameter configurations and time setting. The third one is to test the performance of PN-MCTS in combination with enhancements that are successful in a general-game-playing context such as RAVE \cite{Gelly07} and MAST \cite{bjornsson09}.

\section*{Acknowledgments}

This research is partially funded by the European Research Council as part of the Digital Ludeme Project (ERC Consolidator Grant \#771292).

\bibliographystyle{IEEEtran} 
\bibliography{refs,references} 

\begin{thebibliography}{10}
\providecommand{\url}[1]{#1}
\csname url@samestyle\endcsname
\providecommand{\newblock}{\relax}
\providecommand{\bibinfo}[2]{#2}
\providecommand{\BIBentrySTDinterwordspacing}{\spaceskip=0pt\relax}
\providecommand{\BIBentryALTinterwordstretchfactor}{4}
\providecommand{\BIBentryALTinterwordspacing}{\spaceskip=\fontdimen2\font plus
\BIBentryALTinterwordstretchfactor\fontdimen3\font minus
  \fontdimen4\font\relax}
\providecommand{\BIBforeignlanguage}[2]{{%
\expandafter\ifx\csname l@#1\endcsname\relax
\typeout{** WARNING: IEEEtran.bst: No hyphenation pattern has been}%
\typeout{** loaded for the language `#1'. Using the pattern for}%
\typeout{** the default language instead.}%
\else
\language=\csname l@#1\endcsname
\fi
#2}}
\providecommand{\BIBdecl}{\relax}
\BIBdecl

\bibitem{coulom06}
R.~Coulom, ``Efficient selectivity and backup operators in {M}onte-{C}arlo
  {Tree Search},'' in \emph{Computers and Games (CG 2006)}, ser. Lecture Notes
  in Computer Science, H.~J. van~den Herik, P.~Ciancarini, and H.~H.~L.~M.
  Donkers, Eds., vol. 4630.\hskip 1em plus 0.5em minus 0.4em\relax Berlin
  Heidelberg, Germany: Springer-Verlag, 2007, pp. 72--83.

\bibitem{kocsis06b}
L.~Kocsis and C.~Szepesv\'ari, ``{Bandit Based {Monte-Carlo} Planning},'' in
  \emph{{Machine Learning: ECML 2006}}, ser. Lecture Notes in Artificial
  Intelligence, J.~F\"urnkranz, T.~Scheffer, and M.~Spiliopoulou, Eds., vol.
  4212, 2006, pp. 282--293.

\bibitem{mctssurvey}
C.~B. Browne, E.~Powley, D.~Whitehouse, S.~M. Lucas, P.~I. Cowling,
  P.~Rohlfshagen, S.~Tavener, D.~Perez, S.~Samothrakis, and S.~Colton, ``A
  survey of {M}onte {C}arlo {Tree Search} methods,'' \emph{{IEEE} Transactions
  on Computational Intelligence and {AI} in Games}, vol.~4, no.~1, pp. 1--43,
  2012.

\bibitem{Silver2017mastering}
D.~Silver, J.~Schrittwieser, K.~Simonyan, I.~Antonoglou, A.~Huang, A.~Guez,
  T.~Hubert, L.~Baker, M.~Lai, A.~Bolton, Y.~Chen, T.~Lillicrap, F.~Hui,
  L.~Sifre, G.~van~den Driessche, T.~Graepel, and D.~Hassabis, ``Mastering the
  game of {Go} without human knowledge,'' \emph{Nature}, vol. 550, pp.
  354--359, 2017.

\bibitem{Lorentz08}
R.~J. Lorentz, ``Amazons discover {Monte-Carlo},'' in \emph{Computers and Games
  (CG 2008)}, ser. Lecture Notes in Computer Science, H.~J. van~den Herik,
  X.~Xu, Z.~Ma, and M.~H.~M. Winands, Eds., vol. 5131.\hskip 1em plus 0.5em
  minus 0.4em\relax Berlin Heidelberg, Germany: Springer, 2008, pp. 13--24.

\bibitem{arneson10}
B.~Arneson, R.~B. Hayward, and P.~Henderson, ``{Monte Carlo Tree Search in
  Hex},'' \emph{IEEE Transactions on Computational Intelligence and AI in
  Games}, vol.~2, no.~4, pp. 251--258, 2010.

\bibitem{winands10}
M.~H.~M. Winands, Y.~Bj\"{o}rnsson, and J.-T. Saito, ``{Monte Carlo Tree Search
  in Lines of Action},'' \emph{IEEE Transactions on Computational Intelligence
  and AI in Games}, vol.~2, no.~4, pp. 239--250, 2010.

\bibitem{bjornsson09}
Y.~Bj\"{o}rnsson and H.~Finnsson, ``{CadiaPlayer}: {A} simulation-based
  {General Game Player},'' \emph{IEEE Transactions on Computational
  Intelligence and AI in Games}, vol.~1, no.~1, pp. 4--15, 2009.

\bibitem{sturtevant08}
N.~R. Sturtevant, ``An analysis of {UCT} in multi-player games,'' \emph{ICGA
  Journal}, vol.~31, no.~4, pp. 195--208, 2008.

\bibitem{ciancarini10}
P.~Ciancarini and G.~P. Favini, ``{Monte Carlo Tree Search in Kriegspiel},''
  \emph{AI Journal}, vol. 174, no.~11, pp. 670--–684, 2010.

\bibitem{nijssen12tciaig}
J.~A.~M. Nijssen and M.~H.~M. Winands, ``{Monte-Carlo Tree Search} for the
  hide-and-seek game {Scotland Yard},'' \emph{Transactions on Computational
  Intelligence and AI in Games}, vol.~4, no.~4, pp. 282--294, 2012.

\bibitem{winands08b}
M.~H.~M. Winands, Y.~Bj{\"o}rnsson, and J.-T. Saito, ``{Monte-Carlo Tree Search
  Solver},'' in \emph{Computers and Games (CG 2008)}, ser. Lecture Notes in
  Computer Science (LNCS), H.~J. van~den Herik, X.~Xu, Z.~Ma, and M.~H.~M.
  Winands, Eds., vol. 5131.\hskip 1em plus 0.5em minus 0.4em\relax Berlin
  Heidelberg, Germany: Springer, 2008, pp. 25--36.

\bibitem{winands11}
M.~H.~M. Winands and Y.~Bj\"{o}rnsson, ``$\alpha\beta$-based play-outs in
  {Monte-Carlo Tree Search},'' in \emph{2011 IEEE Conference on Computational
  Intelligence and Games (CIG 2011)}, S.-B. Cho, S.~M. Lucas, and P.~Hingston,
  Eds.\hskip 1em plus 0.5em minus 0.4em\relax IEEE, 2011, pp. 110--117.

\bibitem{LanctotWPS14}
M.~Lanctot, M.~H.~M. Winands, T.~Pepels, and N.~R. Sturtevant, ``Monte carlo
  tree search with heuristic evaluations using implicit minimax backups,'' in
  \emph{2014 {IEEE} Conference on Computational Intelligence and Games, {CIG}
  2014}, 2014, pp. 341--348.

\bibitem{baier2015}
H.~Baier and M.~H.~M. Winands, ``{MCTS}-minimax hybrids,'' \emph{IEEE
  Transactions on Computational Intelligence and AI in Games}, vol.~7, no.~2,
  pp. 167--179, 2015.

\bibitem{allis94}
L.~V. Allis, M.~van~der Meulen, and H.~J. van~den Herik, ``Proof-number
  search,'' \emph{Artificial Intelligence}, vol.~66, no.~1, pp. 91--123, 1994.

\bibitem{vandenHerik2008}
H.~J. van~den Herik and M.~H.~M. Winands, ``Proof-number search and its
  variants,'' in \emph{Oppositional Concepts in Computational Intelligence},
  H.~R. Tizhoosh and M.~Ventresca, Eds.\hskip 1em plus 0.5em minus 0.4em\relax
  Berlin, Heidelberg: Springer Berlin Heidelberg, 2008, pp. 91--118.

\bibitem{breuker}
D.~M. Breuker, ``Memory versus search in games,'' Ph.D. dissertation,
  Maastricht University, Maastricht, The Netherlands, 1998.

\bibitem{nagai}
A.~Nagai, ``Df-pn algorithm for searching {AND/OR} trees and its
  applications,'' Ph.D. dissertation, The University of Tokyo, Tokyo, Japan,
  2002.

\bibitem{Winands04}
M.~H.~M. Winands, J.~W.~H.~M. Uiterwijk, and H.~J. van~den Herik, ``An
  effective two-level proof-number search algorithm,'' \emph{Theoretical
  Computer Science}, vol. 313, no.~3, pp. 511--525, 2004.

\bibitem{Kishimoto05b}
A.~Kishimoto and M.~M\"{u}ller, ``Search versus knowledge for solving life and
  death problems in {G}o,'' in \emph{Proceedings of the 20th National
  Conference on Artificial Intelligence (AAAI'05)}, M.~M. Veloso and
  S.~Kambhampati, Eds.\hskip 1em plus 0.5em minus 0.4em\relax Menlo Park, CA,
  USA: AAAI Press / MIT Press, 2005, pp. 1374--1379.

\bibitem{schaeffer07a}
J.~Schaeffer, N.~Burch, Y.~Bj{\"o}rnsson, A.~Kishimoto, M.~M{\"u}ller, R.~Lake,
  P.~Lu, and S.~Sutphen, ``Checkers is solved,'' \emph{Science}, vol. 317, no.
  5844, pp. 1518--1522, 2007.

\bibitem{Wu10}
I.-C. Wu, H.-H. Lin, P.-H. Lin, D.-J. Sun, Y.-C. Chan, and B.-T. Chen,
  ``Job-level proof-number search for connect6,'' in \emph{Computers and Games
  (CG'10)}, ser. Lecture Notes in Computer Science, H.~J. van~den Herik,
  H.~Iida, and A.~Plaat, Eds., vol. 6515.\hskip 1em plus 0.5em minus
  0.4em\relax Springer-Verlag, Berlin, Germany, 2011, pp. 11--22.

\bibitem{Saito10}
J.-T. Saito and M.~H.~M. Winands, ``Paranoid proof-number search,'' in \emph{In
  Proceedings of the Computational Intelligence and Games Conference (CIG'10)},
  G.~N. Yannakakis and J.~Togelius, Eds.\hskip 1em plus 0.5em minus 0.4em\relax
  IEEE Press, 2010, pp. 203--210.

\bibitem{chaslot08}
G.~M.~J.-B. Chaslot, M.~H.~M. Winands, H.~J. van~den Herik, J.~W.~H.~M.
  Uiterwijk, and B.~Bouzy, ``Progressive strategies for {Monte-Carlo Tree
  Search},'' \emph{New Mathematics and Natural Computation}, vol.~4, no.~3, pp.
  343--357, 2008.

\bibitem{auerfinit02}
P.~Auer, N.~Cesa-Bianchi, and P.~Fischer, ``Finite-time analysis of the
  multiarmed bandit problem,'' \emph{Machine Learning}, vol.~47, no. 2--3, pp.
  235--256, 2002.

\bibitem{Piette2020Ludii}
{\'E}.~Piette, D.~J. N.~J. Soemers, M.~Stephenson, C.~F. Sironi, M.~H.~M.
  Winands, and C.~Browne, ``Ludii -- the ludemic general game system,'' in
  \emph{Proceedings of the 24th European Conference on Artificial Intelligence
  (ECAI 2020)}, ser. Frontiers in Artificial Intelligence and Applications,
  G.~D. Giacomo, A.~Catala, B.~Dilkina, M.~Milano, S.~Barro, A.~Bugarín, and
  J.~Lang, Eds., vol. 325.\hskip 1em plus 0.5em minus 0.4em\relax IOS Press,
  2020, pp. 411--418.

\bibitem{Love_2008_GDL}
N.~Love, T.~Hinrichs, D.~Haley, E.~Schkufza, and M.~Genesereth, ``General game
  playing: Game description language specification,'' Stanford Logic Group,
  Tech. Rep. LG-2006-01, 2008.

\bibitem{sackson69}
S.~Sackson, \emph{A Gamut of Games}.\hskip 1em plus 0.5em minus 0.4em\relax
  Random House, New York, NY, USA, 1969.

\bibitem{awari}
M.~J. Herskovits, ``Wari in the new world.'' \emph{The Journal of the Royal
  Anthropological Institute of Great Britain and Ireland}, vol.~62, p. 23–37,
  1932.

\bibitem{KishimotoW0S12}
A.~Kishimoto, M.~H.~M. Winands, M.~M{\"{u}}ller, and J.-T. Saito, ``Game-tree
  search using proof numbers: The first twenty years,'' \emph{ICGA Journal},
  vol.~35, no.~3, pp. 131--156, 2012.

\bibitem{SilHub18General}
D.~Silver, T.~Hubert, J.~Schrittwieser, I.~Antonoglou, M.~Lai, A.~Guez,
  M.~Lanctot, L.~Sifre, D.~Kumaran, T.~Graepel, T.~Lillicrap, K.~Simonyan, and
  D.~Hassabis, ``A general reinforcement learning algorithm that masters chess,
  shogi, and {Go} through self-play,'' \emph{Science}, vol. 362, no. 6419, pp.
  1140--1144, 2018.

\bibitem{gomoku}
H.~J.~R. Murray, \emph{A History of Board-Games Other Than Chess}.\hskip 1em
  plus 0.5em minus 0.4em\relax Oxford: Clarendon, 1951.

\bibitem{allis96}
L.~V. Allis, H.~J. van~den Herik, and M.~P.~H. Huntjens, ``Go-moku solved by
  new search techniques,'' \emph{Computational Intelligence}, vol.~12, pp.
  7--23, 1996.

\bibitem{TangZSL16}
Z.~Tang, D.~Zhao, K.~Shao, and L.~Lv, ``{ADP} with {MCTS} algorithm for
  {Gomoku},'' in \emph{2016 {IEEE} Symposium Series on Computational
  Intelligence, {SSCI} 2016, Athens, Greece, December 6-9, 2016}.\hskip 1em
  plus 0.5em minus 0.4em\relax {IEEE}, 2016, pp. 1--7.

\bibitem{handscomb01}
K.~Handscomb, ``$8\times$8 game design competition: The winning game:
  {B}reakthrough ... and two other favorites,'' \emph{Abstract Games}, vol.~7,
  pp. 8--9, 2001.

\bibitem{SironiLW20}
C.~F. Sironi, J.~Liu, and M.~H.~M. Winands, ``Self-adaptive {Monte Carlo Tree
  Search} in general game playing,'' \emph{{IEEE} Transactions on Games},
  vol.~12, no.~2, pp. 132--144, 2020.

\bibitem{saffidineJC11}
A.~Saffidine, N.~Jouandeau, and T.~Cazenave, ``Solving {Breakthrough} with race
  patterns and job-level proof number search,'' in \emph{Advances in Computer
  Games - 13th International Conference, {ACG} 2011, Tilburg, The Netherlands,
  November 20-22, 2011, Revised Selected Papers}, ser. Lecture Notes in
  Computer Science, H.~J. van~den Herik and A.~Plaat, Eds., vol. 7168.\hskip
  1em plus 0.5em minus 0.4em\relax Springer, 2011, pp. 196--207.

\bibitem{Gelly07}
S.~Gelly and D.~Silver, ``Combining online and offline knowledge in {UCT},'' in
  \emph{Proceedings of the International Conference on Machine Learning
  (ICML)}, Z.~Ghahramani, Ed.\hskip 1em plus 0.5em minus 0.4em\relax ACM, 2007,
  pp. 273--280.

\end{thebibliography}

\end{document}